\documentclass{article}

\usepackage{microtype}
\usepackage{graphicx}
\usepackage{subcaption}
\usepackage{booktabs}
\usepackage{multirow}

\usepackage[table]{xcolor}
\usepackage{hyperref}

\usepackage[accepted]{icml2026}

\usepackage{amsmath}
\usepackage{amssymb}
\usepackage{mathtools}
\usepackage{amsthm}
\usepackage{algorithm}
\usepackage{algorithmic}
\usepackage{enumitem}

\usepackage[capitalize,noabbrev]{cleveref}

\theoremstyle{plain}

\theoremstyle{definition}

\theoremstyle{remark}

\icmltitlerunning{Is Our Benchmark Enough? An Analysis of Continual Learning for MLLMs}

\begin{document}

\twocolumn[
  \icmltitle{Is Our Benchmark Enough? An Analysis of Continual Learning for MLLMs}

  \icmlsetsymbol{equal}{*}

  \begin{icmlauthorlist}
    \icmlauthor{Van-Tuan Tran}{tcd}
    \icmlauthor{Shruthi Gowda}{tue}
    \icmlauthor{Merim Dzaferagic}{tcd}
    \icmlauthor{Marco Ruffini}{tcd}
  \end{icmlauthorlist}

  \icmlaffiliation{tcd}{School of Computer Science and Statistics, Trinity College Dublin, Ireland}
  \icmlaffiliation{tue}{Department of Mathematics and Computer Science, Eindhoven University of Technology, The Netherlands}

  \icmlcorrespondingauthor{Van-Tuan Tran}{tranva@tcd.ie}

  \icmlkeywords{Continual Adaptation, Mixture of Experts, LoRA,
    Prototype Routing, Parameter-Efficient Fine-Tuning,
    Multimodal Large Language Models, Benchmarking, Sustainable AI}

  \vskip 0.3in
]

\printAffiliationsAndNotice{}

\begin{abstract}
Continual adaptation is essential for multimodal large language models (MLLMs) deployed across evolving domains, but the state-of-the-art MR-LoRA method highly relies on the assumption that a MLLM-based router is necessary to process complex multimodal inputs. This paper revisits this claim on the MLLM-CL benchmark and argues for two claims. \textbf{First}, routing does not require an MLLM: a simple training-free, replay-free ptotypical routing method (\textsc{RePRo}), uses frozen pretrained features and task prototypes to match the MLLM-based router of MR-LoRA at far lower computational cost. \textbf{Second}, shared experts do not improve continual learning for MLLMs, despite their theoretical appeal. We show that these findings arise from two structural limitations of MLLM-CL: (1) its tasks are \textbf{highly separable} in representation space, and (2) its fixed task order makes conclusions \textbf{sensitive to a single curriculum} rather than robust across diverse continual-learning trajectories. As a result, the benchmark primarily rewards learning in isolation rather than genuine continual transfer. This motivates a new design for future benchmarks of continual MLLM learning, with overlapping task manifolds, multiple task orders, fine-grained domain shifts, and evaluation protocols that reward forward transfer as well as retention.

\end{abstract}

\section{Introduction}

Multimodal Large Language Models (MLLMs) \cite{liu2023llava, liu2024improved, chen2024internvl} now serve as the predominant backbone for vision–language applications. Real-world deployment, however, requires ongoing adaptation to new domains, abilities, and data streams under strict computational budgets: retraining from scratch at every update is infeasible at the scale of modern MLLMs, and sequential fine-tuning suffers from catastrophic forgetting on previously learned tasks~\citep{mccloskey1989catastrophic,kirkpatrick2017ewc}. These constraints make \emph{\textbf{continual learning}} (CL) a practical necessity rather than a purely theoretical concern. To support systematic study of this problem, \citet{zhao2025mllmcl} recently introduced MLLM-CL as a comprehensive CL benchmark for MLLMs. 
Within this benchmark, MR-LoRA \citep{zhao2025mllmcl} achieves strong performance by assigning each task an isolated LoRA expert and using an MLLM-based router to select the expert at inference time. While effective, this design raises a fundamental question: \emph{is the MLLM-based router truly necessary, or does the benchmark mainly require identifying which task an input belongs to? }

\begin{table*}[ht]
    \caption{Task accuracy on the Domain Continual Learning of MLLM-CL benchmark. All trainable methods use LoRA rank $r=8$ for fair comparison. Both \textsc{RePro} variants consistently match or exceed the computationally expensive MR-LoRA oracle routing.}
    \label{tab:dcl_performance}
    \vskip 0.15in
    \small
    \begin{center}
    \resizebox{\textwidth}{!}{%
    \begin{tabular}{llcccccc}
    \toprule
    \multirow{2}{*}{\textbf{Method}} & \multirow{2}{*}{\textbf{Details}} & \textbf{Task 1} & \textbf{Task 2} & \textbf{Task 3} & \textbf{Task 4} & \textbf{Task 5} & \textbf{Avg.} \\
    & & \textbf{(RS)} & \textbf{(Med)} & \textbf{(AD)} & \textbf{(Sci)} & \textbf{(Fin)} & \textbf{Acc.} \\
    \midrule
    Zero-shot & No fine-tuning & 32.28 & 28.28 & 15.58 & 43.21 & 62.57 & 36.20 \\
    \midrule
    \multicolumn{8}{l}{\textit{Baselines}} \\
    CL-MoE~\citep{huai2025cl} & MoE of LoRAs with dual momentum router & 69.43 & 34.07 & 31.13 & 38.16 & 86.01 & 51.76 \\
    O-LoRA~\citep{wang2023olora} & Orthogonal subspace LoRA & 21.97 & 28.20 & 14.19 & 35.48 & 58.67 & 31.70 \\
    DISCO-LoRA~\citep{mcitlib2024} & Disentangled shared/task-specific LoRA & 25.50 & 28.75 & 18.70 & 36.04 & 59.65 & 33.73 \\
    HiDe-LoRA~\citep{guo2025hide} & Hierarchical decomposition of LoRA & 28.70 & 26.00 & 18.60 & 36.84 & 52.00 & 32.43 \\
    MoELoRA~\citep{liu2024moelora} & Mixture of LoRA experts with token-level soft routing & 75.23 & 39.79 & 36.61 & 41.47 & 90.83 & 56.79 \\
    SEFE~\citep{chen2025sefe}  & Regularized LoRA against semantic forgetting & 74.38 & 44.60 & 40.24 & 44.41 & 91.47 & 59.02 \\
    Sequential FT & Naive sequential LoRA fine-tuning & 76.70 & 54.60 & 42.65 & 42.71 & 91.85 & 61.70 \\
    EWC~\citep{kirkpatrick2017ewc} & Elastic Weight Consolidation regularization & 77.60 & 44.90 & 32.15 & 40.53 & 90.90 & 57.22 \\
    Experience Replay~\citep{chaudhry2019tinyer} & Rehearsal with stored past samples & 76.50 & 54.75 & 43.85 & 43.67 & 91.65 & 62.08 \\
    MR-LoRA~\citep{zhao2025mllmcl} & Oracle routing, isolated experts & 74.20 & 65.00 & 55.80 & 53.67 & 91.40 & 68.01 \\
    \midrule
    \multicolumn{8}{l}{\textit{Ours}} \\
    \textsc{RePro} & Prototypical routing, Isolated experts, Signal: Vision & 73.80 & 64.60 & 55.80 & 53.67 & 91.40 & 67.85 \\
    \textsc{Multimodal-RePro} & Prototypical Routing, Isolated experts, Signal: Multimodal & 75.60 & 64.70 & 55.80 & 53.77 & 91.40 & \textbf{68.25} \\
    \bottomrule
    \end{tabular}%
    }
    \end{center}
    \vskip -0.1in
\end{table*}

In this work, we take a position by defending two claims. \emph{\textbf{(1) Routing does not require an MLLM}}. We propose Replay-Free Prototypical Routing (\textsc{RePro}), a parameter-free router that uses frozen pretrained representations and task prototypes to select among isolated LoRA experts. Despite its simplicity, \textsc{RePro} matches the MLLM-based router of MR-LoRA while greatly reducing inference cost. This suggests that, on MLLM-CL, expert selection can be solved without MLLM. \emph{\textbf{(2) Shared experts do not improve continual learning for MLLMs.}} Although shared Mixture-of-Experts (MoE) designs are motivated by the idea that related tasks should benefit from shared parameters, our analysis shows that shared experts consistently underperform isolated experts on MLLM-CL. Their training losses converge normally, indicating that the issue is not simply an optimization failure. Rather, the benchmark provides limited transferable structure for shared experts to exploit.

We argue that these two findings reveal \textbf{a broader benchmark limitation}. The tasks in MLLM-CL are highly separable in representation space, making routing nearly trivial, while the use of a single canonical task order makes conclusions sensitive to one curriculum. As a result, the benchmark rewards isolated tasks aptation more than genuine continual transfer. This motivates the next generation of continual MLLM benchmarks: they should include overlapping task manifolds, fine-grained domain shifts, multiple task orders, and evaluation protocols that measure forward transfer as well as retention.

\textbf{Contribution.} We (1) reduce the MR-LoRA routing pipeline to a prototype router that matches its accuracy with eight orders of magnitude fewer FLOPs and three orders lower latency, undermining the assumption that multimodal routing requires an MLLM; (2) report a systematic negative result for shared-expert routing across representative design points, motivating new sharing mechanisms for continual MLLM adaptation; (3) tie both findings to a single cause: the near-linear separability of MLLM-CL tasks (visualized in Fig. \ref{fig:domain_separability}); and (4) outline concrete desiderata for the next generation of continual MLLM benchmarks, including overlapping task manifolds, fine-grained specialization under a shared capability, and evaluation protocols that reward forward transfer rather than merely penalizing forgetting.

\section{Isolated Experts: Are They Really Enough?} \label{sec:analysis}

\subsection{Routing: Does isolation require an MLLM-based router?} \label{sec:isolated-experts}

From Sec.\ref{app:background}, MR-LoRA relies on a heavyweight MLLM-based router tuned on a replay buffer, under the premise that a full MLLM is needed to infer task identity from an image-instruction pair. We challenge this assumption by hypothesizing that if the underlying tasks are conceptually distinct, \emph{their representations in pre-trained foundational spaces are inherently separable} without requiring learned autoregressive routing. To demonstrate this, we propose \textbf{Re}play-Free \textbf{Pro}totypical Routing (\textsc{RePro}), a replay-free, training-free routing methodology, and its \textsc{Multimodal-RePro} variant.

Specifically, after each isolated expert $\theta_k$ is trained on $D_k$, we sample a small prototype support set $S_k$ of $n=128$ examples per task, extract the \texttt{[CLS]} token from the frozen CLIP vision encoder $\phi_v(\cdot)$, and compute an $\ell_2$-normalised centroid:
\begin{equation}
\bar{c}^{v}_k \;=\; \ell_2\!\Bigl(\tfrac{1}{n}\!\sum_{x \in S_k}\!
\ell_2\!\bigl(\phi_v(x)\bigr)\Bigr).
\end{equation}

\begin{table*}[ht]
    \centering
    \caption{Task accuracy and transfer metrics for the proposed \textbf{Mixture of Shared Experts (MoSE)} framework on MLLM-CL benchmark. Despite architectural sophistication and various auxiliary supervisions, shared experts consistently plateau around $\sim$55\%, failing to match the isolated expert baseline ($\sim$68\%).}
    \label{tab:shared_experts_main}
    \resizebox{\textwidth}{!}{
    \begin{tabular}{@{}l l l l ccccc c rr@{}}
    \toprule
    \multirow{2}{*}{\textbf{Routing level}} & \multicolumn{3}{c}{\textbf{Details}} & \textbf{Task 1} & \textbf{Task 2} & \textbf{Task 3} & \textbf{Task 4} & \textbf{Task 5} & \textbf{Avg.} & \multirow{2}{*}{\textbf{BWT}} & \multirow{2}{*}{\textbf{FWT}} \\
    \cmidrule(lr){2-4}
    & \textbf{Gating} & \textbf{Experts} & \textbf{Signal} & \textbf{(RS)} & \textbf{(Med)} & \textbf{(AD)} & \textbf{(Sci)} & \textbf{(Fin)} & \textbf{Acc.} & & \\ \midrule

    \multirow{5}{*}{\begin{tabular}[c]{@{}l@{}}Sample-level\end{tabular}}
     & Softmax, noisy gating & 5 shared experts & Multimodal & 73.70 & 39.70 & 33.90 & 37.73 & 87.70 & \textbf{54.54} & $-7.52$ & 1.34 \\ \addlinespace
     & Softmax, top-$k$ & 5 shared experts, $k=2$ & Multimodal & 72.70 & 34.90 & 21.30 & 35.35 & 84.70 & \textbf{49.79} & $-9.45$ & $-2.27$ \\ \addlinespace
     & Softmax, top-$k$, task-guided & 5 shared experts, $k=2$ & Multimodal & 76.50 & 39.30 & 35.00 & 38.64 & 87.00 & \textbf{55.28} & $-6.52$ & 1.43 \\ \addlinespace
     & \begin{tabular}[c]{@{}l@{}}Softmax, top-$k$, task-guided, \\ orthogonal loss\end{tabular} & 5 shared experts, $k=2$ & Multimodal & 73.70 & 38.40 & 32.00 & 37.73 & 87.70 & \textbf{53.90} & $-8.23$ & 2.88 \\ \midrule

    \multirow{2}{*}{\begin{tabular}[c]{@{}l@{}}Token-level\end{tabular}}
     & Sigmoid & 5 shared experts, soft gating & Multimodal & 75.60 & 38.20 & 33.90 & 38.90 & 88.10 & \textbf{54.94} & $-7.83$ & 0.89 \\ \addlinespace
     & Sigmoid, task-guided & 5 shared experts, soft gating & Multimodal & 76.00 & 39.60 & 35.20 & 38.61 & 87.20 & \textbf{55.32} & $-5.88$ & $-0.80$ \\ \bottomrule
\end{tabular}
}
\end{table*}

At inference, $x$ is routed to $k^\star(x) = \arg\max_k \cos\!\bigl(\phi_v(x), \bar{c}^{v}_k\bigr)$. The \textsc{Multimodal-RePro} variant computes parallel text
prototypes $\bar{c}^{\ell}_k$ from the mean-pooled hidden states of the frozen LLM backbone over the instruction $q$, and fuses modalities by $k^\star(x,q) = \arg\max_k \max\!\bigl(\cos(\phi_v(x), \bar{c}^v_k),
\cos(\phi_\ell(q), \bar{c}^\ell_k)\bigr)$. Detailed algorithmss and additional results are in App. \ref{app:repro_details}.

Tab. \ref{tab:dcl_performance} compares \textsc{RePro} against MR-LoRA's MLLM router and all CL baselines on DCL. \textit{(1) The MLLM router is conceptually redundant.} Replacing it with \textsc{RePro} matches MR-LoRA's accuracy ($67.85$ vs.\ $68.01$), and \textsc{Multimodal-RePro} slightly exceeds it ($68.25$). \textit{(2) Routing cost drops by orders of magnitude.} As shown in Tab. \ref{tab:routing_cost} (App. \ref{app:routing-cost}), MLLM-based router requires $\sim\!1.3\!\times\!10^{13}$~FLOPs and ${\sim}72\,$ms per decision, while \textsc{RePro} operates at $4.1\!\times\!10^{4}$~FLOPs and $\sim\!0.06\,$ms. This means that using a router with \emph{zero} trainable parameters can reduce 8 orders of magnitude in FLOPs and 3 in latency.

These results indicate that reliance on an MLLM router is \textit{an over-engineered solution} to a highly separable representation problem. Future gains under the isolated-expert paradigm must come from improving experts, not routers.

\subsection{Sharing: Can shared experts improve over isolation?} \label{sec:shared-experts}

The foundational premise of Mixture-of-Experts (MoE) and parameter-efficient fine-tuning in continual learning is that shared parameters encode shared structure \citep{dou2024loramoe, mcitlib2024}.
Given that strict isolation explicitly prohibits this parameter-level co-adaptation, we ask a concrete question: \textit{Can we design a shared-expert architecture that extracts cross-task synergies and outperforms strict isolation on MLLM-CL?}

\textbf{MoSE: Mixture of Shared Experts.} \textbf{MoSE} evaluates two representative parameterized-routing paradigms on top of $K\!=\!5$ shared experts. \textit{(1) Sample-level global routing} pools sequence hidden states of both vision and language features, with learnable fusion weights, then applies a noisy softmax gate. \textit{(2) Token-level local routing} operates on individual tokens with a
sigmoid-gated, $\alpha_{t,i} = \sigma\!\bigl(\beta(s_{t,i} - \tau_i)\bigr)$, parameterised by learnable per-expert thresholds $\tau$ and a learnable sharpness $\beta$. We also ablate auxiliary supervisions (task-guided cross-entropy, orthogonality loss) and sparsity constraints (top-$k$). Full algorithmic formulations and initialisation schemes are in App. \ref{app:mose}.

\textbf{Empirical Results and Insights.}
Despite extensive architectural tuning and the theoretical motivation of MoE, \emph{\textbf{no shared-expert
variant outperforms strict isolation}}. Tab. \ref{tab:shared_experts_main} shows that
neither paradigm surpasses the mid-$50\%$ range, with the best shared variant (token-level sigmoid with task supervision) reaching only $55.32\%$. From these results, three diagnostic patterns emerge: \textbf{(1) Hard sparsification drives forgetting} -- sample-level
top-$k$ without task guidance yields the lowest accuracy ($49.79\%$) and the worst BWT ($-9.45\%$); \textbf{(2) Task supervision helps but cannot close the gap} --  task-guided cross-entropy raises sample-level top-$k$ from $49.79\%$ to $55.28\%$ and improves token-level BWT from $-7.83$ to $-5.88$, yet shared architectures remain bottlenecked even when routing is supervised to be
near-perfect; \textbf{(3) Task orthogonality trades retention for transfer} -- adding a task-orthogonality penalty to the top-$k$ task-guided router successfully broke early routing symmetry, yielding the highest FWT in the study ($+2.88\%$), but it hurts retention (BWT drops to $-8.23$). To rule out the possibility that the observed underperformance is simply due to a failure to learn, we report the training-loss curves in Fig. \ref{fig:train-loss-unified} in the Appendix. These curves show that the models converge with nearly identical per-task loss profiles.

\section{The Root Cause Is the Benchmark, Not the Method}
\label{sec:cause}
To understand the underperformance of shared-expert architectures (Sec. \ref{sec:shared-experts}), the surprising efficacy of training-free routing (Sec. \ref{sec:isolated-experts}), we study the properties of MLLM-CL. 
Our analysis reveals that \emph{the failure lies not in the learning algorithms, but in the structural properties of the benchmark itself}. Specifically, we identify two critical artifacts of MLLM-CL that inherently penalize shared parameterization and confound evaluation.

\vspace{-0.25cm}
\subsection{Disjoint Task Manifolds Preclude Knowledge Transfer} \label{sec:task-separation}

The foundation of parameter-efficient sharing in continual learning is that related tasks can synergistically co-activate subsets of parameters  \cite{dou2024loramoe, feng2024mixture, huai2025cl}.  However, these studies assume that the sequential tasks actually exhibit an overlapping representational manifold. 

\begin{figure}[ht]
    \centering
    \includegraphics[width=1\linewidth]{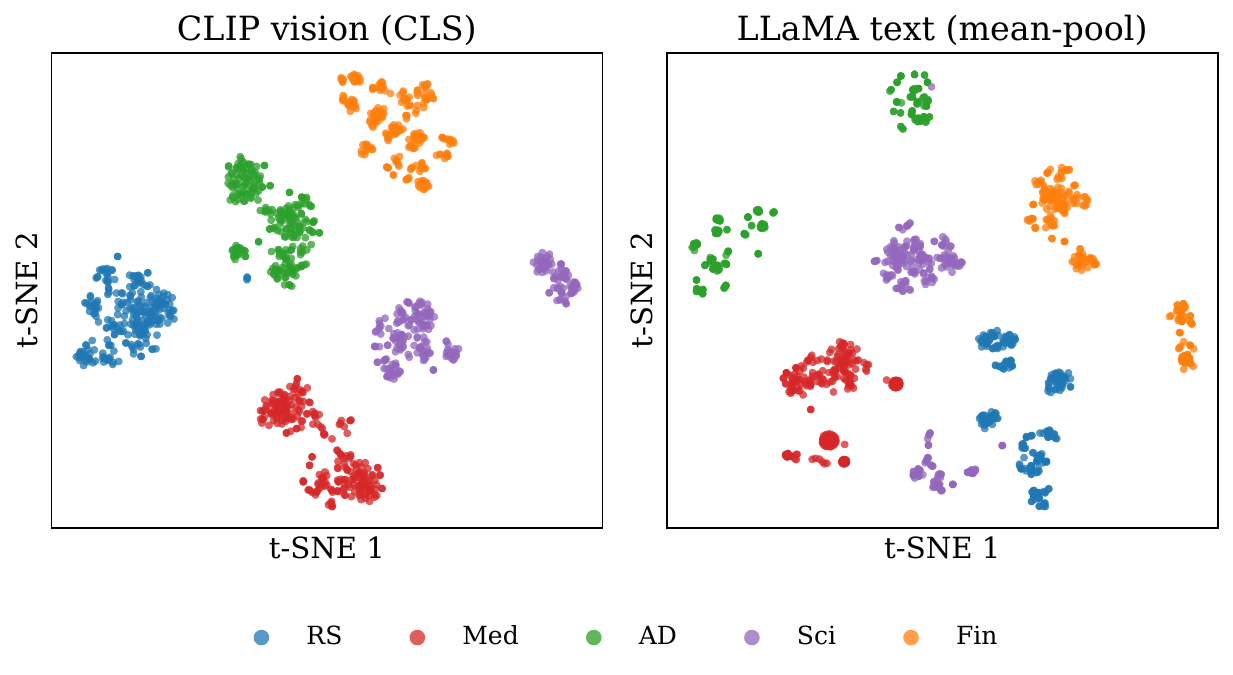}
    \caption{\textbf{The MLLM-CL benchmark is representationally too separate.} t-SNE projections of the five DCL domains under a frozen CLIP visual encoder (left, \texttt{[CLS]} token) and a frozen LLaMA text encoder (right, mean-pooled hidden states). }
    \label{fig:domain_separability}
\end{figure}

As illustrated in the t-SNE projections (Fig. \ref{fig:domain_separability}), this assumption breaks down entirely in the MLLM-CL sequence. When projected into pre-trained foundational spaces, using either a frozen visual encoder (i.e., CLIP \texttt{CLS} tokens) or a textual encoder (i.e., LLaMA~\citep{touvron2023llama} mean-pooled hidden states), 5 DCL domains form perfectly isolated, highly dense clusters. Quantitatively (Tab. \ref{tab:routing_cost} in Appendix), purely textual signals route at $97.4\%$, purely visual at $99.5\%$, and the multimodal router reaches $99.9\%$.

This level of separation has two immediate consequences.
\textbf{(i) The redundancy of complex routing.} Because the representations are cleanly disjoint prior to any task-specific adaptation, a simple router such as \textsc{Multimodal-RePro} can achieve near-oracle expert selection. This explains Sec. \ref{sec:isolated-experts}.
\textbf{(ii) Shared-experts have nothing to share.} Parameter sharing helps only when tasks possess overlapping semantic structure that experts can jointly exploit. 
In this setting, shared low-rank experts are encouraged to fit domain-specific updates rather than reusable cross-domain factors, so later task updates can interfere with parameters needed for earlier tasks instead of producing positive transfer.

\subsection{Task-Order Sensitivity} \label{ssec:task-order}

A second observation is that performance on the canonical DCL sequence is markedly non-uniform across positions. In both Tabs. \ref{tab:dcl_performance}--\ref{tab:shared_experts_main}, every method peaks on Task 1 (RS) and Task 5 (Fin) and bottoms out on the interior positions Tasks 2–4 (Med, AD, Sci). We hypothesize that this order affects the performance of shared experts.

To evaluate our hypothesis, we re-evaluated shared-expert architectures across multiple task permutations. To ensure these permutations are principled, we anchor them on the inherent difficulty of each domain, quantified by the zero-shot accuracy of the unadapted base model (LLaVA-1.5~\citep{liu2024improved}). This establishes a strict difficulty gradient from the hardest to the easiest domain: Autonomous Driving (AD, 15.58\%) $<$ Medical (Med, 28.28\%) $<$ Remote Sensing (RS, 32.28\%) $<$ Science (Sci, 42.31\%) $<$ Finance (Fin, 62.57\%). Across the complete four-order ablation, all evaluated shared-pool MoE tuners follow the same order ranking: \texttt{easy-to-hard} $>$ \texttt{random} $>$ \texttt{canonical} $\geq$ \texttt{hard-to-easy} (Details in Tab. \ref{tab:task_orders}). As shown in Tab. \ref{tab:order_spread}, the resulting order spread is large (6.36--8.21 pp) and exceeds the largest between-method spread within any fixed order (4.73 pp), showing that the DCL order can dominate the choice of gating function.

\begin{table}[ht]
    \centering
    \caption{Mean accuracy (\%) across the five DCL domains, $\bar{a}_5(m, \pi)$, evaluated after the final task under each task order. $\Delta_{\text{order}}$ reports the spread between the best and worst order for each method. Best per method in \textbf{bold}.}
    \label{tab:order_spread}
    \small
    \begin{tabular}{@{}lccc@{}}
    \toprule
    \textbf{Task order} & \textbf{CL-MoE} & \textbf{MoELoRA} & \textbf{MoSE-sigmoid} \\
    \midrule
    \texttt{canonical} & 51.77 & 47.04 & 49.70 \\
    \texttt{hard-to-easy} & 48.92 & 46.15 & 47.81 \\
    \texttt{easy-to-hard} & \textbf{55.28} & \textbf{54.36} & \textbf{55.04} \\
    \texttt{random} & 50.52 & 48.98 & 51.20 \\
    \midrule
    $\Delta_{\text{order}}$ & 6.36 & 8.21 & 7.23 \\
    \bottomrule
    \end{tabular}
\end{table}

\section{Desiderata for the Next CL Benchmark for MLLMs}
\label{sec:desiderata}

Our results establish three constraints for the next continual MLLM benchmark. \textbf{First}, trivial routing must not be the hidden solution: on MLLM-CL, frozen CLIP/LLaMA features route domains with more than $99\%$ accuracy, and a nearest-prototype router matches an MLLM router. \textbf{Second}, sharing cannot be judged on tasks with no shared manifold: all shared-expert MoE variants converge normally yet plateau far below isolated experts. \textbf{Third}, a single canonical order is insufficient: changing the DCL order drastically affects shared-expert MoE accuracy, even exceeding the difference between gating designs.
 
A benchmark should therefore formalize the hard part of CL: sparse, sequential exposure to shifted data, without reducing the problem to simultaneous multi-task training or large-scale replay. 
This requires streams where task identity is not linearly recoverable from frozen representations, but where tasks still share enough visual concepts, instructions, output formats, or latent rules for positive transfer to be possible. It should include recurring and compositional tasks, distinguish domain shift from capability shift, and evaluate multiple task orders calibrated by the zero-shot base model.
Strong methods should retain old skills, acquire new ones, selectively overwrite stale knowledge, and improve on related future tasks without explicit task labels, unbounded expert growth, or full replay.

\section{Conclusion}
\label{sec:conclusion}

This paper shows that continual learning on MLLM-CL can be solved more simply than current routing-heavy designs suggest. \textsc{RePro} replaces the MLLM-based router with replay-free prototype routing over frozen features, achieving comparable performance at far lower cost. We also find that shared experts fail to outperform isolated experts, not because they cannot learn, but because the benchmark provides limited transferable structure. Together, these results reveal two limitations of MLLM-CL: highly separable task representations and reliance on a single task order. Future benchmarks should include overlapping task manifolds, multiple curricula, and metrics that reward genuine transfer as well as retention.

\newpage
\section*{Acknowledgements}
The ``MoSE: Mixture-of-Specialized-Experts for efficient continual adaptation" action has received funding from the European Union, via the oc3-2025-TES-01 issued and implemented by the ENFIELD project, under the grant agreement No 101120657.

\bibliography{references}
\bibliographystyle{icml2026}

\newpage
\appendix

\section{Background} \label{app:background}

This section establishes the formal setting and the technical building blocks used throughout the paper. We first define continual learning for MLLMs, then describe the MLLM-CL benchmark that all our experiments use, and finally summarize the MR-LoRA pipeline that our analysis takes as its reference design.

\subsection{Continual Learning for MLLMs}
\label{sec:cl-formulation}

Following \citet{zhao2025mllmcl}, an MLLM is a function $f_\theta$ that maps an image--instruction pair $(x^{\text{img}}, x^{\text{ins}})$ to an answer $y$ of $L$ tokens, trained autoregressively. In the continual setting the model observes a sequence of $T$ tasks $\{\mathcal{D}_t\}_{t=1}^{T}$, where each $\mathcal{D}_t = \{(x^{\text{img}}_{t,i}, x^{\text{ins}}_{t,i}, y_{t,i})\}_{i=1}^{N_t}$ is drawn IID from a task-specific distribution $\mathcal{P}_t = \mathcal{X}^{\text{img}}_t \times \mathcal{X}^{\text{ins}}_t \times \mathcal{Y}_t$. At stage $t$ only $\mathcal{D}_t$ is available; data from tasks $1, \dots, t-1$ cannot be revisited at full scale. The training loss at stage $t$ is the standard next-token objective,
\begin{equation}
\mathcal{L}_t(\theta) = -\sum_{i=1}^{N_t} \sum_{l=1}^{L} \log p_\theta\!\left(y^{l}_{t,i} \mid x^{\text{img}}_{t,i},\, x^{\text{ins}}_{t,i},\, y^{<l}_{t,i}\right).
\label{eq:cl-loss}
\end{equation}
After the full sequence has been seen, the model is evaluated on test inputs drawn from any of $\{\mathcal{P}_j\}_{j=1}^{T}$. The two failure modes of interest are \emph{catastrophic forgetting}, the degradation on tasks $1, \dots, t-1$ after training on $\mathcal{D}_t$, and \emph{loss of plasticity}, the failure to acquire $\mathcal{D}_t$ itself due to interference with prior parameters. Performance is reported with the standard CL metrics: Average accuracy after the final task, forward transfer (FWT), and backward transfer (BWT).

\subsection{MLLM-CL Benchmark} \label{sec:mllm-cl}

Several vision--language continual-learning benchmarks precede MLLM-CL, but they target different model scales and problem formulations. CLiMB~\citep{srinivasan2022climb} evaluates multimodal task-to-task learning and downstream low-shot transfer with ViLT-style encoders and classification-oriented heads. VQACL~\citep{zhang2023vqacl} focuses on VQA continual learning, organizing tasks by question type and visual category to test novel skill--concept compositions. CoIN~\citep{chen2024coin} moves toward continual instruction tuning for MLLMs, but it only benchmarks the forgetting aspect of MLLM methods across a chain of instruction-tuning datasets. We therefore use MLLM-CL~\citep{zhao2025mllmcl} as the single testbed for this paper because it best matches our target setting: benchmarking MLLM's ability to obtain new domains/abilities and standardized continual-learning metrics for evaluation.

MLLM-CL comprises two regimes: Domain Continual Learning (DCL), which measures absorption of domain-specific knowledge under IID train/test splits, and Ability Continual Learning (ACL), which targets fundamental capabilities (OCR, math \& logic, visual perception, GUI agent) under non-IID splits.

We restrict our analysis to DCL because two ACL tasks lack deterministic ground truth: math \& logic labels are produced by an LLM-as-judge, and GUI agent answers are scored by an AI-based grading platform. Both inject a stochastic, model-driven component into the evaluation, which would confound the controlled comparisons our position paper claims rely on.

DCL comprises five tasks: remote sensing (RS), medical (Med), autonomous driving (AD), science (Sci), and finance (Fin) — presented in the fixed order:

\begin{equation*}
\text{RS} \;\to\; \text{Med} \;\to\; \text{AD} \;\to\; \text{Sci} \;\to\; \text{Fin},
\label{eq:dcl-order}
\end{equation*}
with roughly 60k training and 10k test examples per task, IID within each task.

\subsection{MR-LoRA: Isolated Experts with an MLLM Router}
\label{sec:mr-lora}

MR-LoRA \citep{zhao2025mllmcl} is built on Low-Rank Adaptation (LoRA;~\citealp{hu2022lora}), which freezes a pretrained weight $W_0$ and adds a low-rank update $\Delta W = BA$ with $r \ll \min(d, k)$, training only $A$ and $B$. MR-LoRA combines LoRA with strict per-task isolation and a generative router.

\paragraph{Isolated expert training.}
For each task $t$, MR-LoRA trains a dedicated LoRA $\phi_t$ on $\mathcal{D}_t$ from the frozen MLLM backbone. Experts are mutually non-interacting: $\phi_t$ never sees $\mathcal{D}_{<t}$, and $\phi_{<t}$ are never updated when $\mathcal{D}_t$ arrives. By construction, this eliminates parameter-level forgetting, since the past experts are bit-identical at the end of training as they were when frozen. The cost is that some mechanism must decide, at inference, \emph{which} expert to apply to a given input.

\paragraph{MLLM-based routing.}
MR-LoRA treats expert selection as a generation task. After each stage, it collects a small replay buffer $\mathcal{M}_t = \{(x^{\text{img}}_{t,i}, x^{\text{ins}}_{t,i})\}_{i=1}^{m}$ with $m=20 \ll N_t$ and finetunes a separate \emph{router LoRA} $\phi_R$ on $\bigcup_{j \le t} \mathcal{M}_j$ using a structured prompt that lists all expert descriptions. Given a test input, the full MLLM with $\phi_R$ runs a forward pass over the prompt and image and autoregressively emits a single token: the identifier of the chosen expert. The selected $\phi_{\hat{t}}$ is then swapped in to produce the final answer in a second forward pass.

Two properties of this design matter for the rest of the paper. First, every test query incurs \emph{two} MLLM forward passes, one to route and one to answer, even though the routing pass produces only one token. Second, the experts are entirely decoupled at training time: the architecture admits no cross-task gradient flow and no parameter sharing beyond the frozen backbone.

\section{Related Work}
\label{app:related}

\paragraph{CL for MLLMs.} CoIN~\cite{chen2024coin} and
MLLM-CL~\cite{zhao2025mllmcl} established continual instruction tuning
benchmarks for MLLMs. 
Comes with MLLM-CL is the MR-LoRA baseline, which allocates one fresh LoRA expert per task to avoid parameter interference. 
The price of this isolation is a replay-trained Router LoRA that runs the full MLLM to generate an expert identifier for every test query.
Other multimodal adaptation methods use different forms of expert or
representation management: MoCLE~\cite{gou2023mocle} clusters instructions offline and activates cluster-conditional LoRA experts with a universal expert for generalization; D-MoLE~\cite{ge2025dynamic} dynamically allocates layer-wise LoRA experts and uses a modality curriculum for continual multimodal instruction tuning; and C-CLIP~\cite{liu2025cclip} studies continual learning for CLIP-style image--text matching while preserving zero-shot ability. 

\paragraph{MoE-LoRA with learned routers.}
LoRA-based MoE and adaptive-adapter families include LoRAMoE~\cite{dou2024loramoe}, MoELoRA~\cite{liu2024moelora}, MixLoRA~\cite{li2024mixlora}, Mixture-of-LoRAs~\cite{feng2024mixture}, Uni-MoE~\cite{li2025unimoe}, SiRA~\cite{zhu2023sira}, MoLE~\cite{wu2024mole}, MoA~\cite{cao2025moa}, HMoRA~\cite{liao2025hmora}, AdaMoLE~\cite{liu2024adamole}, AM-LoRA~\cite{liu2024amlora}, and Online-LoRA~\cite{wei2025onlinelora}.
Most MoE-style variants assume a fixed expert pool and train a softmax, top-$k$, attentional, or
thresholded router, often with load-balancing, contrastive, or curriculum to prevent expert collapse. 
TT-LoRA-MoE~\cite{kunwar2025tt} is architecturally closest to the isolated-expert side of our analysis because it trains specialized low-rank experts independently and then freezes them; however, its
router is a learned top-1 gate, the task set is predefined before router training, and its evaluation is text-only rather than continual MLLM routing.

\paragraph{Isolation, orthogonality, and prototypes.}
Parameter-level non-interference is enforced by O-LoRA~\cite{wang2023olora}, InfLoRA~\cite{liang2024inflora}, CL-LoRA~\cite{he2025cllora}, SD-LoRA~\cite{wu2025sdlora}, and LoRA-Subtraction~\cite{lorasub2025}. 
They constrain new updates through orthogonal subspaces, null-space projections, shared/task-specific adapter splits, direction--magnitude decoupling, or drift-resistant spaces. 
This line of work addresses forgetting by shaping where LoRA updates may live, whereas expert isolation style avoids interference by never sharing task-specific updates and then delegating the remaining problem to a router. Prototype-based CL traces back to iCaRL~\cite{rebuffi2017icarl}, where class means support incremental recognition. Our contribution is to lift this idea from class prototypes to task prototypes over frozen visual and language representations, and to use those prototypes as a replay-free, zero-parameter router for generative MLLM experts.

\paragraph{MLLM-CL baselines.} The baselines in Tab.~\ref{tab:dcl_performance} cover the main families of continual learning. \emph{Replay/regularization.} EWC~\cite{kirkpatrick2017ewc} adds a quadratic Fisher-weighted penalty that anchors parameters important to past tasks, while Experience Replay~\cite{chaudhry2019tinyer} interleaves stored past samples with new-task batches. SEFE~\cite{chen2025sefe} applies element-wise regularization to LoRA updates so that parameters consolidated for past tasks are protected while new-task adaptation continues. \emph{Subspace isolation.} O-LoRA~\cite{wang2023olora} constrains each new task's LoRA update to a subspace orthogonal to those of all previous tasks. \emph{MoE-LoRA routing.} MoELoRA~\cite{liu2024moelora} attaches a token-level soft router over a fixed pool of LoRA experts trained with a contrastive load-balancing objective; CL-MoE~\cite{huai2025cl} extends this with dual momentum updates on the router and experts to trade off plasticity and stability for continual VQA. \emph{Hierarchical/structured LoRA for MLLMs.} HiDe-LoRA~\cite{guo2025hide} hierarchically decouples task-shared and task-specific LoRA components for continual instruction tuning of MLLMs, and DISCO-LoRA (as implemented in MCITlib~\cite{mcitlib2024}) keeps a per-task LoRA bank and selects parameter embeddings at inference via textual similarity to a codebook of past instructions. \emph{Isolated experts with a learned MLLM router.} MR-LoRA~\cite{zhao2025mllmcl} trains one fresh LoRA expert per task in isolation and uses a replay-tuned MLLM-LoRA that autoregressively emits the expert identifier at inference.

\section{Algorithmic Details and Additional Results for Prototypical Routing (\textsc{RePro})}
\label{app:repro_details}

This appendix provides the training configuration, prototype bank construction protocol, and expanded empirical ablations for the \textsc{RePro} router introduced in Sec.~\ref{sec:isolated-experts}. We restate the routing pipeline in Sec.~\ref{app:repro:pipeline}, document the isolated expert-training regime in Sec.~\ref{app:repro:isolation}, describe prototype bank construction in Sec.~\ref{app:repro:prototypes}, and report additional diagnostics in Sec.~\ref{app:repro:results}.

\subsection{\textsc{RePro} and \textsc{Multimodal-RePro}}
\label{app:repro:pipeline}

\textsc{RePro} decouples expert selection from autoregressive answer generation. The pipeline consists of three stages:

\begin{enumerate}[leftmargin=*, nosep]
    \item \textbf{Isolated expert training.} For each domain $k \in [K]$, we estimate a task-specific LoRA adapter $\theta_k$ by minimizing the autoregressive negative log-likelihood on $\mathcal{D}_k$:
    \begin{equation}
        \mathcal{L}_{\text{LM}}(\theta_k) = - \sum_{(x, q, y) \in \mathcal{D}_k} \sum_{t=1}^{|y|} \log p_{\theta_k}(y_t \mid x, q, y_{<t}).
    \end{equation}
    The backbone, previous adapters, and routing rule are held fixed, so expert training introduces no router-specific gradients.

    \item \textbf{Prototype bank construction.} For each domain $k$, we sample a prototype support set $\mathcal{S}_k \subset \mathcal{D}_k$ with $|\mathcal{S}_k|=n$ ($n=128$ in our experiments). We extract the \texttt{[CLS]} representation from the frozen CLIP vision encoder, $\phi_{v}(\cdot)$, and compute an $\ell_2$-normalized centroid:
    \begin{equation}
        \bar{\mathbf{c}}^v_k = \ell_2\!\left(\frac{1}{n}\sum_{x \in \mathcal{S}_k} \ell_2\big(\phi_{v}(x)\big)\right).
        \label{eq:centroid}
    \end{equation}

    \item \textbf{Nearest-prototype routing.} At inference time, a query image $x$ is assigned to the expert whose prototype has maximum cosine similarity to the frozen visual representation:
    \begin{equation}
        k^\star(x) = \arg\max_{k \in [K]}\ \ell_2\big(\phi_{v}(x)\big) \cdot \bar{\mathbf{c}}^v_k.
        \label{eq:routing}
    \end{equation}
\end{enumerate}

The \textsc{Multimodal-RePro} variant adds instruction-level task evidence. It constructs text prototypes $\bar{\mathbf{c}}^\ell_k$ from mean-pooled hidden states of the frozen LLM backbone, $\phi_{\ell}(q)$, and applies late fusion:
\begin{equation}
    \resizebox{1\linewidth}{!}{%
        $k^\star(x, q) = \arg\max_{k \in [K]} \max\!\Big(\cos\big(\phi_{v}(x), \bar{\mathbf{c}}^v_k\big),\ \cos\big(\phi_{\ell}(q), \bar{\mathbf{c}}^\ell_k\big)\Big)$
    }
    \label{eq:max_fusion}
\end{equation}
The complete procedure is provided in Alg.~\ref{alg:repro}.

\begin{algorithm}[t]
    \caption{\textsc{RePro}: Replay-Free Prototypical Routing}
    \label{alg:repro}
    \begin{algorithmic}[1]
    \REQUIRE base MLLM $f_\theta$, frozen vision encoder $\phi_v$,
             frozen text encoder $\phi_\ell$, task stream
             $\{\mathcal{D}_k\}_{k=1}^{K}$, support-set size $n$
    \STATE \textbf{Stage 1: Isolated expert training}
    \FOR{$k = 1, \ldots, K$}
       \STATE Load $f_\theta$ with a fresh LoRA $\theta_k$
       \STATE Optimize $\theta_k$ on $\mathcal{D}_k$ using $\mathcal{L}_{\text{LM}}$
    \ENDFOR
    \STATE \textbf{Stage 2: Prototype bank construction}
    \FOR{$k = 1, \ldots, K$}
       \STATE Sample prototype support set $\mathcal{S}_k \subset \mathcal{D}_k$, $|\mathcal{S}_k|=n$
       \STATE $\bar c_k^{v} \leftarrow \ell_2\!\big(\tfrac{1}{n}\!\sum_{(x,q,y)\in\mathcal{S}_k}\!\ell_2(\phi_v(x))\big)$
       \STATE \textit{(\textsc{Multimodal-RePro})} $\bar c_k^{\ell} \leftarrow \ell_2\!\big(\tfrac{1}{n}\!\sum_{(x,q,y)\in\mathcal{S}_k}\!\ell_2(\phi_\ell(q))\big)$
    \ENDFOR
    \STATE \textbf{Stage 3: Nearest-prototype routing} \hfill (per query $(x,q)$)
    \STATE $k^\star \leftarrow \arg\max_k \cos\!\big(\phi_v(x), \bar c_k^{v}\big)$
    \STATE \textit{(\textsc{Multimodal-RePro})} $\;k^\star \leftarrow \arg\max_k \max\!\big(\cos(\phi_v(x), \bar c_k^v),\,\cos(\phi_\ell(q), \bar c_k^\ell)\big)$
    \STATE Activate expert $\theta_{k^\star}$ in $f_\theta$ and decode the answer
    \end{algorithmic}
\end{algorithm}

\subsection{Expert Isolation and Training Configurations}
\label{app:repro:isolation}

To implement strict expert isolation, each task $T_k$ in the continual-learning sequence is trained in an independent DeepSpeed process. We load the base LLaVA-1.5 model, attach a fresh rank-8 LoRA adapter, and optimize only that adapter on the corresponding domain dataset $\mathcal{D}_k$. No trainable parameters are shared across task-specific adapters.

\textbf{Hyperparameters.} Training is conducted for $1$ epoch per task using a learning rate of $10^{-4}$ with a cosine decay schedule, $0.03$ warmup ratio, per-device batch size of $4$, and gradient accumulation of $2$. No LoRA weights, optimizer states, or scheduler momentum are carried over between tasks.

\subsection{Prototype Bank Construction}
\label{app:repro:prototypes}

After training the $K=5$ isolated experts, we construct a prototype bank from prototype support sets $\mathcal{S}_k$ with $128$ training instances per domain $k \in \{\text{RS, Med, AD, Sci, Fin}\}$. Let $x_i^{(k)}$ denote the image and $q_i^{(k)}$ denote the human-turn instruction. Answers are excluded to prevent label leakage into textual prototypes. We evaluate three feature spaces:
\begin{enumerate}[leftmargin=*, nosep]
    \item \textbf{Vision ($\phi_v$):} The \texttt{[CLS]} token of the frozen LLaVA vision feature extractor ($d=1024$).
    \item \textbf{Text-CLIP ($\phi_t$):} The text pooler output from the frozen CLIP text encoder ($d=768$).
    \item \textbf{Text-LLM ($\phi_\ell$):} The mean-pooled last-hidden state of the frozen LLaMA backbone over non-pad tokens ($d=4096$). By operating on tokenized text directly, we bypass multimodal label preparation, requiring no image tokens for this step.
\end{enumerate}
For each modality $m \in \{v, t, \ell\}$, the domain prototype is the $\ell_2$-normalized mean of the $\ell_2$-normalized feature vectors. Prototype bank construction for all five domains requires approximately 3 minutes on a single H100 GPU.

\subsection{Additional Results}
\label{app:repro:results}

\subsubsection{Per-Modality Routing Accuracy}
\label{app:repro:modality}

\begin{table}[ht]
    \caption{\textbf{Prototypical Routing Accuracy by Modality.} Evaluated over a 5{,}000-sample intersection test set using the \textsc{RePro} router. Near-perfect routing accuracies of unimodal signals confirm the extreme linear separability of MLLM-CL task domains in the latent space, rendering over-parameterized routers redundant.}
    \label{tab:repro_modality}
    \vskip 0.1in
    \begin{center}
    \begin{small}
    \begin{sc}
    \setlength{\tabcolsep}{3pt}
    \begin{tabular}{llcc}
    \toprule
    Modality & Encoder & Route & Avg. Task \\
     &  & Acc. & Acc. \\
    \midrule
    Text       & LLaMA        & 97.42 & 67.65 \\
    Vision     & CLIP & 99.52 & 68.13 \\
    Multimodal & CLIP + LLaMA    & 99.88 & 68.23 \\
    \bottomrule
    \end{tabular}
    \end{sc}
    \end{small}
    \end{center}
    \vskip -0.1in
\end{table}

A single modality is sufficient: pure CLIP vision already routes at
$99.52\%$, and pure LLaMA text at $97.42\%$. Adding the second modality gets an additional $0.36$ percentage points of routing accuracy and only $0.10$ percentage points of task accuracy. The signal is consistent with the t-SNE visualisation in Fig. \ref{fig:domain_separability}: domains are pre-separated in both pretrained spaces, so the routing problem reduces to nearest-centroid lookup.

\subsubsection{Modality Fusion Ablation}
To rigorously determine the most effective mechanism for combining
$\phi_v$ and $\phi_\ell$ in \textsc{Multimodal-RePro}, we conducted an
ablation over different fusion rules. Let $s_k \in \mathbb{R}$
denote the routing score for expert~$k$. The principal configurations
are:
\begin{itemize}
  \item \textbf{Weighted Sum}: $s_k = \alpha\cos(\phi_v,\bar{\mathbf{c}}^v_k) + (1-\alpha)\cos(\phi_\ell,\bar{\mathbf{c}}^\ell_k)$,
  with $\alpha\in\{0.0, 0.2, 0.4, 0.5, 0.6, 0.8, 1.0\}$. Pure text is
  $\alpha=0$; pure vision is $\alpha=1$.
  \item \textbf{Concatenation}: $s_k = \cos\!\big(\ell_2(\phi_v\oplus\phi_\ell),\,\ell_2(\bar{\mathbf{c}}^v_k\oplus\bar{\mathbf{c}}^\ell_k)\big)$.
  \item \textbf{Z-Norm Sum}: $s_k = z(\cos^v_k) + z(\cos^\ell_k)$, with
  per-sample $z$-score over $k$, correcting for scale mismatches between
  the $1024$-d and $4096$-d spaces.
  \item \textbf{Max Fusion}: $s_k = \max\!\big(\cos(\phi_v,\bar{\mathbf{c}}^v_k),\,\cos(\phi_\ell,\bar{\mathbf{c}}^\ell_k)\big)$.
\end{itemize}

In this ablation, fusion rules (\textbf{Concatenation}) and statistical normalizations (\textbf{Z-norm Sum}) fail to outperform naive late-fusion. \textbf{Max Fusion} proves to be the optimal fusion rule. Notably, CLIP-Text prototypes ($\phi_t$) were empirically redundant due to cross-modal contrastive alignment during pre-training; therefore, we use LLaMA hidden states ($\phi_\ell$) for the textual modality.

\subsubsection{Routing Cost}
\label{app:routing-cost}

\begin{table}[ht]
    \caption{Measured routing cost on H100 NVL, batch size of 1, datatype is bf16. $L=32, T=64, d=4096, d_\phi=1024, K=5$. The MR-LoRA routing prompt is $999$ tokens (Vicuna system prompt + the routing instructions and expert descriptions of \citet{zhao2025mllmcl} + 576 image tokens + the user question), so we report the full per-query routing pass: CLIP-ViT-L/14-336 forward, the projector head, and a prefill process of 999 tokens followed by a single decoded token.}
    \label{tab:routing_cost}
    \vskip 0.15in
    \begin{center}
    \begin{sc}
    \resizebox{\columnwidth}{!}{%
    \begin{tabular}{lcccc}
    \toprule
    Router & Params  & FLOPs & ms \\
    \midrule
    Oracle Routing (MR-LoRA, full) & 7.06 B  & $1.33\times 10^{13}$ & 71.82 \\
    \quad CLIP-ViT-L/14-336 & 0.30 B & $3.5\times 10^{11}$ & 5.78 \\
    \quad mm\_projector & 21 M & $2.4\times 10^{10}$ & 0.10 \\
    \quad Llama-LM (prefill 999 + 1) & 6.74 B & $1.30\times 10^{13}$ & 65.93 \\
    Token Softmax (CL-MoE) & 20 K & $8.4\times 10^{7}$ & 53.10 \\
    \textbf{Ours (\textsc{RePro})} & \textbf{0} & $\mathbf{4.1\times 10^{4}}$ & \textbf{0.06} \\
    \bottomrule
    \end{tabular}%
    }
    \end{sc}
    \end{center}
    \vskip -0.1in
\end{table}

\textsc{RePro} reduces the per-decision cost by approximately
\emph{eight orders of magnitude in FLOPs} and \emph{three orders of
magnitude in latency} relative to MR-LoRA's MLLM-LoRA selector. The router has zero learnable parameters and a $20$~KB on-disk footprint for the prototype bank, so it adds no gradient interference, no optimiser state, and no GPU memory overhead at training time. Combined with the prototype bank construction cost of $\sim\!3$ minutes on a single GPU, the entire routing pipeline, including training, prototype bank construction, and inference, is dominated by the cost of the experts themselves.

\section{Algorithmic Details and Extended Results for Mixture of Shared Experts (MoSE)}
\label{appendix:shared-experts}
\label{app:mose}

In this section, we provide the algorithmic formulations, auxiliary objective details, and extended empirical results for the \textbf{Mixture of Shared Experts (MoSE)} framework discussed in Section~4.2. All shared-expert models use a fixed expert cardinality of $K=5$, matching the number of task domains in the DCL sequence. We use the same descriptive variant names as the main text rather than internal run identifiers.

\subsection{Sample-level Global Routing}

Sample-level global routing is a parameterized shared-expert router that computes one routing distribution per input sequence and applies that distribution uniformly to all tokens. The design is intended to capture coarse task identity from pooled multimodal representations before expert composition.

\textbf{Algorithmic Formulation.} Given an input sequence of hidden states $\{x_1, x_2, \dots, x_T\}$ with $x_t \in \mathbb{R}^d$, we first pool the sequence to obtain a global representation $\bar{x} \in \mathbb{R}^d$:
\begin{equation}
    \bar{x} = \frac{1}{T}\sum_{t=1}^{T} x_t
\end{equation}
The router features two parallel branches—a linear vision branch and a non-linear language multi-layer perceptron (MLP) branch:
\begin{equation}
    s_v = W_v \bar{x}, \qquad s_\ell = W_\ell^{(2)} \operatorname{ReLU}\!\big(W_\ell^{(1)} \bar{x}\big)
\end{equation}
where $W_v \in \mathbb{R}^{K \times d}$. The language projection bottleneck dimension is $256$, so $W_\ell^{(1)} \in \mathbb{R}^{256 \times d}$ and $W_\ell^{(2)} \in \mathbb{R}^{K \times 256}$. The modalities are dynamically fused via learnable branch weights $w \in \mathbb{R}^2$ (initialized to $[0.5, 0.5]$):
\begin{equation}
    \hat{w} = \operatorname{softmax}(w), \qquad s = \hat{w}_1 s_v + \hat{w}_2 s_\ell
\end{equation}
During training, we inject Gaussian noise to encourage expert exploration: $s' = s + \varepsilon \sigma_n$ where $\varepsilon \sim \mathcal{N}(0, 1)$ and $\sigma_n=0.1$. The final routing mixture is $r = \operatorname{softmax}(s')$. For every token $t$, the layer output is computed as:
\begin{equation}
    y_t = W_0 x_t + \frac{\alpha}{r_{lora}} \sum_{k=1}^{K} r_k \big(B_k A_k x_t\big)
\end{equation}
where $r_{lora}$ is the LoRA rank. This dual-branch architecture adds $\sim$1M parameters per target module.

\textbf{Ablation configurations and analysis.} We evaluate four sample-level configurations in Table~\ref{tab:shared_experts_main}:
\begin{itemize}
    \item \textbf{Noisy softmax routing.} The router is trained only through the language-modeling loss $\mathcal{L}_{LM}$, with Gaussian routing noise for exploration. Router parameters are initialized from $\mathcal{N}(0, 0.01^2)$.
    \item \textbf{Top-$k$ sparsified softmax routing.} The router retains only the largest $k=2$ logits and masks the remaining logits to $-\infty$ before the softmax. This hard sparsification produces premature commitment to early domains and yields the worst backward transfer in the sample-level family (BWT $=-9.45\%$).
    \item \textbf{Top-$k$ routing with task-guided supervision.} We add cross-entropy supervision on the routing logits, $\mathcal{L}_{tg} = \operatorname{CE}(\operatorname{softmax}(s), k_{gt})$, with $\lambda_{tg}=1.0$. This improves task differentiation and gives the best sample-level mean accuracy ($55.28\%$), but it remains well below isolated expert routing.
    \item \textbf{Top-$k$ routing with task-guided supervision and orthogonal loss.} We use a wider initialization, $\mathcal{N}(0, 0.1^2)$, and penalize expert overlap with a task-similarity-weighted orthogonality objective, $\mathcal{L}_{orth} = \sum_{i<j} (1 - \operatorname{sim}(T_i, T_j)) \|A_i A_j^\top\|_F^2$. This increases forward transfer (FWT $=+2.88\%$) but reduces retention, consistent with the trade-off reported in the main text.
\end{itemize}

\subsection{Token-level Local Routing}

Token-level local routing tests whether the shared-expert plateau is specific to sample-level assignment. Instead of assigning one mixture to the whole sequence, this router computes token-specific expert weights using a continuous, differentiable sigmoid gate.

\textbf{Algorithmic Formulation.} For each individual token $t$, the dense hidden state $x_t$ is linearly projected to base routing logits $s_t = W_r x_t \in \mathbb{R}^K$. To enable independent suppression or activation of experts without hard masking, we compute a per-token, per-expert gate $\alpha_{t,i} \in (0,1)$ using a parameterized sigmoid function:
\begin{equation}
    \alpha_{t,i} = \sigma\!\Big(\beta\,(s_{t,i} - \tau_i)\Big)
\end{equation}
where $\tau \in \mathbb{R}^K$ are learnable per-expert thresholds (initialized to $0$) and $\beta \in \mathbb{R}$ is a learnable global scaling factor controlling the sharpness of the threshold (initialized to $1.0$). The final mixture weights for the token are computed via a masked softmax:
\begin{equation}
    r_t = \operatorname{softmax}(s_t \odot \alpha_t)
\end{equation}
This parameter-efficient dense router requires only $\sim$20K parameters per module.

\textbf{Ablation configurations and analysis.} We evaluate two token-level configurations in Table~\ref{tab:shared_experts_main}:
\begin{itemize}
    \item \textbf{Sigmoid routing.} The router is trained through $\mathcal{L}_{LM}$ without explicit task labels and obtains $54.94\%$ mean accuracy. Empirical logs show that $\beta$ increases to approximately $1.5$--$2.0$ and the thresholds $\tau$ diverge around zero, indicating that the gate learns non-trivial expert modulation rather than collapsing to a uniform mixture.
    \item \textbf{Sigmoid routing with task-guided supervision.} We add sequence-aggregated task supervision, $\mathcal{L}_{tg} = \operatorname{CE}(\operatorname{softmax}(\bar{s}), k_{gt})$ with $\bar{s}=\frac{1}{T}\sum_{t=1}^{T}s_t$ and $\lambda_{tg}=0.1$. This improves mean accuracy to $55.32\%$ and reduces backward forgetting (BWT improves from $-7.83\%$ to $-5.88\%$), but the result still falls far short of isolated expert routing.
\end{itemize}

\subsection{Training-Loss Diagnostics}
\label{app:loss-curves}

\begin{figure*}[t]
    \centering
    \includegraphics[width=\textwidth]{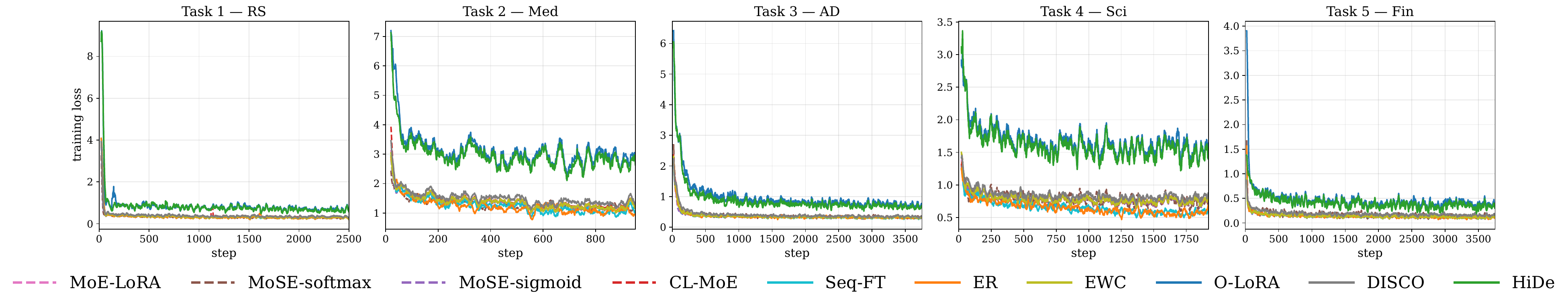}
    \caption{\textbf{Shared-MoE models optimize normally despite poor final accuracy.} Training-loss curves across the five DCL tasks. MoSE-softmax, MoSE-sigmoid, MoELoRA, Seq-FT, ER, and EWC exhibit similar convergence patterns within each task, ruling out failed optimization as the primary explanation for the shared-expert performance plateau. O-LoRA and HiDe-LoRA are the main exceptions: their auxiliary objectives keep the reported training loss elevated on several tasks, consistent with their weak DCL accuracy.}
    \label{fig:train-loss-unified}
\end{figure*}

\cref{fig:train-loss-unified} provides an optimization diagnostic for the
negative shared-expert result. If MoSE failed because the shared routers or
experts simply did not learn, its training loss should diverge, plateau early,
or show a qualitatively worse decay profile than the sequential and replay
baselines. Instead, the loss curves for MoSE-softmax and MoSE-sigmoid closely
track the standard baselines on each task. The final logged losses for the
full-data MoE-style runs are also nearly matched: MoELoRA, MoSE-softmax, and
MoSE-sigmoid end at 0.354/0.347/0.352 on RS, 1.332/1.317/1.294 on Med,
0.786/0.760/0.775 on Sci, and 0.145/0.148/0.146 on Fin. Where all three are
available for AD, MoSE-softmax and MoSE-sigmoid also end near Seq-FT
(0.352/0.363 vs.\ 0.352 for Seq-FT with rank 32).

This evidence rules out a simple optimization-failure account. \textbf{The shared-MoE
models do learn the current task objective, but their learned updates still do
not transfer into high retained accuracy across the DCL sequence}. The failure is
therefore better explained by cross-task interference and the absence of useful
overlap in the task manifolds, as argued in Sec. \ref{sec:cause}.

The exceptions are informative. O-LoRA and HiDe-LoRA include auxiliary
regularization/decomposition terms that change the reported training objective
rather than optimizing the language-modeling loss alone. Their curves remain
substantially higher on several domains, especially Med and Sci, and these are
also among the weakest methods in Tab. \ref{tab:dcl_performance}. Thus, unlike
MoSE and MoELoRA, their poor benchmark performance is at least partly
consistent with objective-level optimization pressure from the auxiliary terms.

\begin{table}[ht]
    \centering
    \caption{Task orders evaluated in the order-sensitivity ablation. Difficulty is anchored to the zero-shot accuracy of the un-adapted backbone on each domain.}
    \label{tab:task_orders}
    \small
    \setlength{\tabcolsep}{4pt}
    \begin{tabular}{@{}lp{2.5cm}p{3cm}@{}}
    \toprule
    \textbf{Order} & \textbf{Sequence} & \textbf{Rationale} \\
    \midrule
    \texttt{canonical} & RS $\to$ Med $\to$ AD $\to$ Sci $\to$ Fin & Original order from the MLLM-CL benchmark. \\
    \addlinespace[2pt]
    \texttt{hard-to-easy} & AD $\to$ Med $\to$ RS $\to$ Sci $\to$ Fin & Anti-curriculum: hardest first. \\
    \addlinespace[2pt]
    \texttt{easy-to-hard} & Fin $\to$ Sci $\to$ RS $\to$ Med $\to$ AD & Curriculum: easiest first. \\
    \addlinespace[2pt]
    \texttt{random}    & Sci $\to$ Fin $\to$ AD $\to$ RS $\to$ Med & Random permutation \\
    \bottomrule
    \end{tabular}
\end{table}

\begin{table*}[ht]
    \centering
    \vspace{-0.8em}
    \caption{Final accuracy (\%) under task-order ablations. Can.=canonical, H$\rightarrow$E=hard-to-easy, E$\rightarrow$H=easy-to-hard.}
    \label{tab:order_breakdown}
    \setlength{\tabcolsep}{3.2pt}
    \renewcommand{\arraystretch}{0.88}
    \begin{tabular}{@{}llrrrrrr@{}}
    \toprule
    \textbf{Method} & \textbf{Order} & \textbf{RS} & \textbf{Med} & \textbf{AD} & \textbf{Sci} & \textbf{Fin} & \textbf{Mean} \\
    \midrule
    CL-MoE       & Can.             & 69.42 & 34.08 & 31.12 & 38.25 & 86.00 & 51.77 \\
                 & H$\rightarrow$E  & 65.65 & 31.70 & 24.45 & 37.22 & 85.60 & 48.92 \\
                 & E$\rightarrow$H  & 69.00 & 33.25 & 46.75 & 43.24 & 84.15 & 55.28 \\
                 & Rand.            & 63.45 & 34.30 & 32.15 & 37.15 & 85.55 & 50.52 \\
    \midrule
    MoELoRA      & Can.             & 57.45 & 30.55 & 27.85 & 36.93 & 82.40 & 47.04 \\
                 & H$\rightarrow$E  & 62.40 & 30.70 & 18.35 & 36.16 & 83.15 & 46.15 \\
                 & E$\rightarrow$H  & 66.20 & 35.75 & 45.05 & 42.50 & 82.30 & 54.36 \\
                 & Rand.            & 59.00 & 31.95 & 32.65 & 36.64 & 84.65 & 48.98 \\
    \midrule
    MoSE-sigmoid & Can.             & 68.15 & 31.20 & 26.30 & 37.17 & 85.70 & 49.70 \\
                 & H$\rightarrow$E  & 63.45 & 31.05 & 20.65 & 37.63 & 86.25 & 47.81 \\
                 & E$\rightarrow$H  & 67.85 & 35.10 & 45.70 & 41.63 & 84.90 & 55.04 \\
                 & Rand.            & 65.15 & 34.80 & 33.00 & 37.10 & 85.95 & 51.20 \\
    \bottomrule
    \end{tabular}
    \vspace{-1.0em}
\end{table*}

\subsection{Task-Order Sensitivity Ablation}
\label{app:task-order}

The main DCL protocol evaluates all methods on the canonical MLLM-CL order,
RS $\to$ Med $\to$ AD $\to$ Sci $\to$ Fin. For shared-pool MoE tuners, this
single order can confound method quality with the position at which each domain
is encountered: early domains are exposed to all subsequent overwriting updates,
while late domains are evaluated shortly after training. We therefore repeated
the DCL protocol under the four permutations in \cref{tab:task_orders} and
measured
\begin{equation}
\Delta_{\text{order}}(m)
= \max_{\pi \in \Pi} \bar{a}_5(m,\pi)
- \min_{\pi \in \Pi} \bar{a}_5(m,\pi),
\end{equation}
where $\bar{a}_5(m,\pi)$ is the mean final-task accuracy across RS, Med, AD,
Sci, and Fin after method $m$ is trained under order $\pi$.

All task-order runs use LLaVA-1.5-7B with the CLIP-ViT-L/14-336 as the vision feature extractor, $E=5$ LoRA experts, LoRA rank $r=30$ ($r/E=6$ per expert for split-rank
CoIN-style MoE variants), LoRA $\alpha=60$, one epoch per task, AdamW with a
$10^{-4}$ learning rate and cosine schedule, a uniform training fraction
$\rho=0.1$, and an evaluation cap of $1{,}000$ examples per domain. Sequential
training warm-starts task $t$ from the adapter saved after task $t-1$ and then
evaluates the final task-5 adapter on all five domains.

\paragraph{Insights.}
\cref{tab:order_spread} shows that task order dominates the gating function:
all three evaluated methods follow the same ranking,
\texttt{easy-to-hard} $>$ \texttt{random} $>$ \texttt{canonical} $\geq$
\texttt{hard-to-easy}. The within-method order spread is 6.36--8.21 pp, while the
largest between-method spread within a fixed order is only 4.73 pp. Thus, for
shared-pool MoE tuners on DCL, which permutation is used matters more than
whether the router is top-$k$ noisy softmax, dense softmax, or dense sigmoid.

The main driver is late-task forgetting of the hardest domain. AD has the
lowest zero-shot accuracy (15.58\%) and the largest order-induced swing:
CL-MoE, MoELoRA, and MoSE-sigmoid reach 46.75\%, 45.05\%, and 45.70\% when AD
is task 5 in \texttt{easy-to-hard}, but only 24.45\%, 18.35\%, and 20.65\% when AD
is task 1 in \texttt{hard-to-easy}. In contrast, Finance, the easiest domain
(62.57\% zero-shot), varies by less than 4 pp across all completed orders.
Canonical DCL is therefore a near-worst case at $\rho=0.1$: it is the worst
order for MoELoRA and the second-worst order for CL-MoE and MoSE-sigmoid.

MoSE-softmax is omitted from \cref{tab:order_spread,tab:order_breakdown}
because it shares the same CoIN-style softmax forward pass and parameter shape
as MoELoRA; the task-order orchestrators intentionally skip it as a parity
duplicate rather than substitute full-fraction legacy numbers into the
$\rho=0.1$ matrix.

\end{document}